\documentclass{article}

\usepackage{url}

\usepackage{graphicx}

\usepackage[english]{babel}

\usepackage{amsmath}
\usepackage{amssymb}
\usepackage{wasysym}
\usepackage[colorlinks=true, allcolors=blue]{hyperref}

\usepackage[linesnumbered,ruled]{algorithm2e}

\usepackage{tikz}
\usetikzlibrary{fit,positioning,calc,matrix,shapes.multipart,shapes.misc,backgrounds,math}

\newcommand{\FF}{\mathcal{F}}
\newcommand{\xx}{\mathbf{x}}
\newcommand{\XX}{\mathbf{X}}
\newcommand{\yy}{\mathbf{y}}
\newcommand{\YY}{\mathbf{Y}}
\newcommand{\ww}{\mathbf{w}}
\newcommand{\WW}{\mathbf{W}}
\newcommand{\WSWS}{\mathsf{WS}}
\newcommand{\ZSZS}{\mathsf{ZS}}
\newcommand{\WAWA}{\mathsf{WA}}
\newcommand{\HSHS}{\mathsf{HS}}
\newcommand{\HAHA}{\mathsf{HA}}
\newcommand{\ypad}{\mathsf{y\_pad}}
\newcommand{\wpad}{\mathsf{w\_pad}}
\newcommand{\xpad}{\mathsf{x\_pad}}

\makeatletter
\def\endthebibliography{%
  \def\@noitemerr{\@latex@warning{Empty `thebibliography' environment}}%
  \endlist
}
\makeatother

\title{Invertible Convolution with Symmetric Paddings}
\author{Bo Li\\
\texttt{\footnotesize prclibo@gmail.com}}
\date{}

\begin{document}
\maketitle

\begin{abstract}
We show that symmetrically padded convolution can be analytically inverted via DFT. We comprehensively analyze several different symmetric and anti-symmetric padding modes and show that multiple cases exist where the inversion can be achieved. The implementation is available at \url{https://github.com/prclibo/iconv_dft}.
\end{abstract}

\section{Introduction}

Convolutional Neural Network (CNN) has achieved incredible success in computer vision. It is shown to be effective in building transformation between the highly complex distribution of image data domain to a latent feature domain distribution that is easy to conduct tasks like classification or generation. Recent studies have shown that there exist CNN operators that are invertible. This makes it possible to build bidirectional transformation between the image data domain and the latent feature domain. Such transformation is recognized to be valuable in unifying the formulation of generative and discriminative tasks and provides potential insights in revealing the interior process of CNNs.

A plenty of research has been conducted in the past decades in designing invertible CNN operators. As in Section~\ref{sec:related-works}, these works mainly focuses two directions: 1) designing CNN operators with special structures that can be analytically inverted; 2) constraining CNN training so that the network can be numerically inverted. Besides these two approaches, it is known but less studied that the vanilla convolution structure is analytically invertible if satisfying very simple conditions. One typical example is that circularly-padded convolution with equal input and output channels~\cite{finzi2019invertible}. Compared with other approaches, such structure provides most similar structural and behavioral analogy to conventional CNN and supports analytical inversion.

We noticed that a disadvantage of the circular convolution is that circular padding violates the spatial locality of CNN and might confuse the training of the high layers whose input feature map sizes are small. A latter research suggests that convolution with symmetric padding and symmetric kernel is also invertible. However, existing study on the symmetric padding case is still limited and only focuses on the simplest case. In this paper, we provide a comprehensive study on the invertibility of symmetric padded convolution. We show that multiple invertible convolution operators exist with different combination of the symmetric padding modes of the input and the kernel. With our analysis we are able to convert conventional a CNN network to its invertible counterpart with only very small modification of structure and efficiently compute analytical inversion with DFT.

\section{Related Works}
\label{sec:related-works}

Normalizing flows design several special convolution operators to allow inversion computation. These approaches are able to exactly compute likelihood of the data and have attracted wide attention in recent research. 
Representative works of normalizing flows include NICE, RealNVP and Glow. These works 
replace conventional convolution layer with the specifically designed coupling layers that are analytically invertible. Compared with conventional convolution layers, the coupling layers are less flexible as the its convolution is only applied on half on the input features. To increase flexibility, a $1\times1$ invertible convolution is appended to a coupling layer. A variety of works have been proposed to further extend the methodology of normalizing flow. For example, \cite{huang2020augmented, chen2020vflow, grcic2021densely} make use of extra noise to augment the feature space and increase capacity. \cite{grcic2021densely} also enhances the performance by introducing more complex layer connection.

The invertibility can also be achieved if the convolution follows certain auto-regressive formulation. \cite{hoogeboom2019emerging} replace a conventional convolution kernel with an auto-regressive variant which only operates the prefix pixels at a given position. \cite{ma2019macow} improves auto-regressive kernels to by rotating masked kernels and claims better efficiency. These design is also shown to be effective in constructing normalizing flows.

Coupling layers and auto-regressive structures both modifies conventional convolutions. Recently there also have been some study on the invertibility of conventional convolutions. By interpreting convolution
in the DFT domain, it is early known that convolution can be inverted by division in the frequency domain. \cite{karami2018generative, finzi2019invertible} discuss this property in the circularly padded convolution. \cite{karami2019symmetric} explores a simple case of symmetric padding which could enable invertible convolution. 

Besides the above analytical approach to invert a convolution operation. Various works have studied to invert CNN in a numerical manner. \cite{behrmann2019invertible} is a representative work which makes use of fixed-point iteration approach to numerically solve the network input from the output.
\cite{grathwohl2019your} does not directly invert a CNN network from a given output, but makes it possible to draw samples from the input image domain given a energy-based network.

Although invertible networks have attracted wide-spread interest in machine learning research in the past decades, it has to be admitted that the performance of invertible networks is still disadvantageous compared to conventional discriminative and generative models. More theoretical development is still required to push the understanding of invertible networks and the bidirectional relationship between image data domain and the latent feature domain. The derivation in this paper is a further development to the study of~\cite{karami2018generative, karami2019symmetric, finzi2019invertible}. 

\section{Methodologies}

\subsection{Preliminaries}

In the derivation of this section, we first consider 1D signal cases for simplicity and then extend it to 2D cases. Define $\xx = x_0, x_1, \dots, x_{N - 1}$ as a 1D discrete signal of length $N$ and $\XX = X_0, X_1, \dots, X_{N - 1}$ as its DFT as
\begin{equation}
    \XX = \FF(\xx), \quad X_k = \sum_{n = 0}^{N - 1} x_n \cdot \exp(- \frac{i 2 \pi}{N} k n).
    \label{eq:dft}
\end{equation}
Obviously, the convolution operation between a kernel $\ww$ and $\xx$ with circular padding corresponds to multiplication in the frequency domain of $\XX$ and the DFT $\WW$ of $\ww$:
\begin{equation}
    \yy = \mathsf{circular\_conv}(\ww, \xx) \longleftrightarrow \YY = \WW \circ \XX,
    \label{eq:conv-dft}
\end{equation}
where $y$ is the output signal and $Y$ is its DFT.

\subsection{Invertibility of Circularly Padded Convolution}

One can recover $\yy$ from $\xx$ and $\ww$ via the Fourier domain by
\begin{equation}
    \XX = \mathsf{element\_inverse}(\WW) \circ \YY.
    \label{eq:invert-freq}
\end{equation}
In practical computer vision models, $\ww$ and $\WW$ are 4D tensors with dimensions of height, width, input channels and output channels. $\mathsf{element\_inverse}$ denotes element-wise matrix inversion on each matrix element in $\WW$ with the dimensions of input channels and output channels. Two conditions are required for this inversion.
\begin{enumerate}
    \item $\WW$ elements are invertible, i.e., input and output channels equal and the element matrices are non-singular.
    \item $\xx$ and $\yy$ should be fully known, i.e., if $\xx$ is padded, values in $\yy$ that corresponds to the padding area can be inferred.
\end{enumerate}
In the practice of deep learning, if a signal is arbitrarily padded, the output feature map is actually a crop of the convolved padded signal. Without knowing the cropped area of the output feature map, it is generally not possible to compute $\YY$.

One special case is the circular padded convolution. In this cases, the cropped area of the output feature map is periodical repeat of the output feature map. Thus $\YY$ can be directly obtained as the DFT of the output feature map.

We next show that besides circular padding, there exist multiple symmetric padding modes that also support the inversion of convolution from the output feature maps.

\subsection{Symmetric Padding}
\label{sec:symmetric-padding}

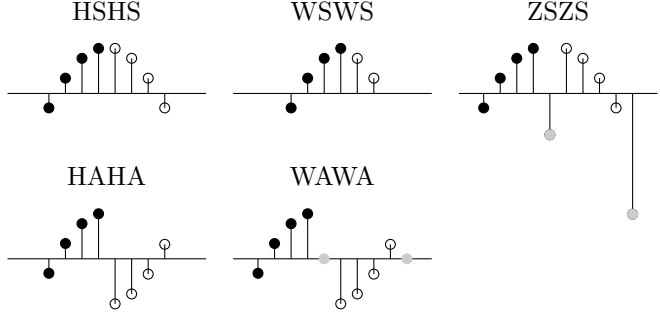
\begin{figure}
    \centering
    \begin{tikzpicture}

% https://tex.stackexchange.com/questions/123760/draw-crosses-in-tikz
\tikzset{cross/.style={cross out, draw, 
         minimum size=2*(#1-\pgflinewidth), 
         inner sep=0pt, outer sep=0pt}}

\tikzset{
    hshs/.pic={
        \node at (0, 5) {HSHS};
        \draw (-6, 0) -- (6, 0);
        \foreach \i in {-3.5,...,-0.5}{
            \tikzmath{\x = \i; \y = -0.3 * \x * \x + 2.8;}
            \draw[] (\x,0) -- (\x, \y);
            \draw[fill] (\x, \y) circle(0.3);
        }
        \foreach \i in {0.5,...,3.5}{
            \tikzmath{\x = \i; \xn = -\x; \y = -0.3 * \xn * \xn + 2.8;}
            \draw[] (\x,0) -- (\x,\y);
            \draw[] (\x,\y) circle(0.3);
        }
    }
}

\tikzset{
    wsws/.pic={
        \node at (0, 5) {WSWS};
        \draw (-6, 0) -- (6, 0);
        \foreach \i in {-2.5,...,0.5}{
            \tikzmath{\x = \i; \xn = \x - 1; \y = -0.3 * \xn * \xn + 2.8;}
            \draw[] (\x,0) -- (\x, \y);
            \draw[fill] (\x, \y) circle(0.3);
        }
        \foreach \i in {1.5,...,2.5}{
            \tikzmath{\x = \i; \xn = -\x; \y = -0.3 * \xn * \xn + 2.8;}
            \draw[] (\x,0) -- (\x,\y);
            \draw[] (\x,\y) circle(0.3);
        }
    }
}

\tikzset{
    zszs/.pic={
        \node at (0, 5) {ZSZS};
        \draw (-6, 0) -- (6, 0);
        \foreach \i in {-4.5,...,-1.5}{
            \tikzmath{\x = \i; \xn = \x + 1; \y = -0.3 * \xn * \xn + 2.8;}
            \draw[] (\x,0) -- (\x, \y);
            \draw[fill] (\x, \y) circle(0.3);
        }
        \tikzmath{\x = -4.5; \xn = \x + 1; \ya = -0.3 * \xn * \xn + 2.8;};
        \tikzmath{\x = -2.5; \xn = \x + 1; \yb = -0.3 * \xn * \xn + 2.8;};
        \tikzmath{\y = -2 * (\ya + \yb);};
        \draw (-0.5, 0) -- (-0.5, \y) circle(0.3);
        \draw [gray!40, fill=gray!40] (-0.5, \y) circle(0.3);
        \foreach \i in {0.5, ..., 3.5}{
            \tikzmath{\x = \i; \xn = -\x; \y = -0.3 * \xn * \xn + 2.8;}
            \draw[] (\x,0) -- (\x,\y);
            \draw[] (\x,\y) circle(0.3);
        }
        \tikzmath{\x = -3.5; \xn = \x + 1; \ya = -0.3 * \xn * \xn + 2.8;};
        \tikzmath{\x = -1.5; \xn = \x + 1; \yb = -0.3 * \xn * \xn + 2.8;};
        \tikzmath{\y = -2 * (\ya + \yb);};
        \draw (4.5, 0) -- (4.5, \y) circle(0.3);
        \draw [gray!40, fill=gray!40] (4.5, \y) circle(0.3);
    }
}

\tikzset{
    haha/.pic={
        \node at (0, 5) {HAHA};
        \draw (-6, 0) -- (6, 0);
        \foreach \i in {-3.5,...,-0.5}{
            \tikzmath{\x = \i; \xn = \x; \y = -0.3 * \xn * \xn + 2.8;}
            \draw[] (\x,0) -- (\x, \y);
            \draw[fill] (\x, \y) circle(0.3);
        }
        \foreach \i in {0.5,...,3.5}{
            \tikzmath{\x = \i; \xn = -\x; \yn = -0.3 * \xn * \xn + 2.8; \y = -\yn;}
            \draw[] (\x,0) -- (\x,\y);
            \draw[] (\x,\y) circle(0.3);
        }
    }
}

\tikzset{
    wawa/.pic={
        \node at (0, 5) {WAWA};
        \draw (-6, 0) -- (6, 0);
        \foreach \i in {-4.5,...,-1.5}{
            \tikzmath{\x = \i; \xn = \x + 1; \y = -0.3 * \xn * \xn + 2.8;}
            \draw[] (\x,0) -- (\x, \y);
            \draw[fill] (\x, \y) circle(0.3);
        }
        \draw [gray!40, fill=gray!40] (-0.5, 0) circle(0.3);
        \foreach \i in {0.5, ..., 3.5}{
            \tikzmath{\x = \i; \xn = -\x; \yn = -0.3 * \xn * \xn + 2.8; \y = -\yn;}
            \draw[] (\x,0) -- (\x,\y);
            \draw[] (\x,\y) circle(0.3);
        }
        \draw [gray!40, fill=gray!40] (4.5, 0) circle(0.3);
    }
}
\pic at (0, 0) [scale=0.22] {hshs};
\pic at (3, 0) [scale=0.22] {wsws};
\pic at (6, 0) [scale=0.22] {zszs};
\pic at (0, -2.2) [scale=0.22] {haha};
\pic at (3, -2.2) [scale=0.22] {wawa};

\end{tikzpicture} 
    \caption{1D Illustration of several symmetric padding modes mentioned in this paper. The black dots denotes a 1D discrete original signal whose length is $N = 4$. Empty circles denotes the reflection of the original signal in the padding area. Gray dots denots extra added data samples to achieve some required properties on the padded signal, c.f. Section~\ref{sec:symmetric-padding}. Note that the signal is periodical and the left padding is wrapped to the right side.}
    \label{fig:pad-mode}
\end{figure}

As in~\cite{martucci1994symmetric}, 4 different padding modes exist for symmetric padding, namely half-sample symmetry (HS), whole-sample symmetry (WS), half-sample anti-symmetry (HA) and whole-sample anti-symmetry (WA).  

Fig.~\ref{fig:pad-mode} shows an example of the 4 padding modes in a 1D signal case.
HS reflects the whole signal. WS keeps the boundary point as the axis of symmetry and reflect the rest of the signal.
HA rotates the signal $180^\circ$ around a half-sample at the boundary. WA rotates the signal $180^\circ$ around an extra zero sample next to the signal boundary.

For simplicity, we assume that all sides of an image input are always padded in the same mode.
A 1D signal that is padded in HS mode on both sides is denoted as padded in HSHS mode. We denote the padded signal as $\HSHS(\xx)$ and its DFT as $\XX^\HSHS$. Similar notation goes with other padding modes.
Symmetric padding flip the signal and append it to the origin one. We reveal several properties of symmetrically padded signals.
\begin{itemize}
    \item In our case where the padding modes are the same on both sides, $N$ of the padded signal is always even.
    % \item For signals padded in HSHS and WSWS, their DFT are pure real, i.e., $\textrm{real}(X_i) = 0$.
    \item $\WAWA(\xx)$ and $\HAHA(\xx)$ have zero summation, i.e., $X^\WAWA_0 = \sum_n x^\WAWA_n = 0$ and $X^\HAHA_0 = \sum_n x^\HAHA_n = 0$.
    \item $\HSHS(\xx)$ and $\WAWA(\xx)$ satisfies $X^\HSHS_{N / 2} = \sum_n (-1)^n x^\HSHS_n = 0$ and $X^\WAWA_{N / 2} = \sum_n (-1)^n x^\WAWA = 0$.
\end{itemize}

For an original signal with even length $N$, we further define an extra padding mode zero-summed padding (ZSZS). For a signal $x$, we reflect the signal and put extra samples of $2 \sum_{n=0}^{N/2} x_{2 n}$ and $2 \sum_{n=0}^{N/2} x_{2 n + 1}$ respectively on the two sides between the origin and reflected signal. Obviously,
\begin{itemize}
    \item ZSZS is a special case of WSWS. ZSZS pads the signal similarly on its two sides, i.e., by adding an extra sample.
    \item Signals padded in ZSZS have zero summation ($X_0$). 
    \item  In addition, substituting into \eqref{eq:dft} reveals that the ZSZS-padded signal also satisfies $X_{N / 2} = 0$.
\end{itemize}

We next give a rough illustration on the invertibility of the symmetrically padded convolution.

\begin{table}[]
    \setlength{\tabcolsep}{2pt}

    \centering
    \begin{tabular}{r|c|cc|c|cc|c|cc|c}
         & $x$ & $X_0$ & $X_{N/2}$ & $w$ & $W_0$ & $W_{N/2}$ & $y$ & $Y_0$ & $Y_{N/2}$ & Inv\\\hline
         1 & HAHA & $0$ & $*$ & WSWS & $*$ & $*$ & HAHA & $0$ & $*$ & \checkmark\\
         2 & WAWA & $0$ & $0$ & WSWS & $*$ & $*$ & WAWA & $0$ & $0$ & \checkmark\\
         3 & HSHS & $*$ & $0$ & WSWS & $*$ & $*$ & HSHS & $*$ & $0$ & \checkmark \\
         4 & WSWS & $*$ & $*$ & WSWS & $*$ & $*$ & WSWS & $*$ & $*$ & \checkmark\\
         5 & ZSZS & $0$ & $0$ & WSWS & $*$ & $*$ & ZSZS & $0$ & $0$ & \checkmark\\\hline
         6 & HAHA & $0$ & $*$ & HSHS & $*$ & $0$ & WAWA & $0$ & $0$ & \\
         7 & WAWA & $0$ & $0$ & HSHS & $*$ & $0$ & HAHA & $0$ & $0$ & \\
         8 & HSHS & $*$ & $0$ & HSHS & $*$ & $0$ & WSWS& $*$ & $0$  &  \\
         9 & WSWS & $*$ & $*$ & HSHS & $*$ & $0$ & HSHS & $*$ & $0$ & \\
         10 & ZSZS & $0$ & $0$ & HSHS & $*$ & $0$ & HSHS & $0$ & $0$ & \\\hline
         11 & HAHA & $0$ & $*$ & WAWA & $0$ & $0$ & HSHS & $0$ & $0$ & \\
         12 & WAWA & $0$ & $0$ & WAWA & $0$ & $0$ & ZSZS & $0$ & $0$ & \checkmark \\
         13 & HSHS & $*$ & $0$ & WAWA & $0$ & $0$ & HAHA & $0$ & $0$ & \\
         14 & WSWS & $*$ & $*$ & WAWA & $0$ & $0$ & WAWA & $0$ & $0$ & \\
         15 & ZSZS & $0$ & $0$ & WAWA & $0$ & $0$ & WAWA & $0$ & $0$ & \checkmark\\\hline
         16 & HAHA & $0$ & $*$ & HAHA & $0$ & $*$ & WSWS & $0$ & $*$ & \\
         17 & WAWA & $0$ & $0$ & HAHA & $0$ & $*$ & HSHS & $0$ & $0$ & \\
         18 & HSHS & $*$ & $0$ & HAHA & $0$ & $*$ & WAWA & $0$ & $0$ &  \\
         19 & WSWS & $*$ & $*$ & HAHA & $0$ & $*$ & HAHA & $0$ & $*$ & \\
         20 & ZSZS & $0$ & $0$ & HAHA & $0$ & $*$ & HAHA & $0$ & $0$ & \\\hline
    \end{tabular}
    \caption{Padding mode transition in symmetrically padded convolution}
    \label{tab:transition}
\end{table}

\subsection{Invertibility}

We restrict our discussion for original signals with even length $N$ since we will make use of the ZSZS mode to help analyze the transition relationship of padding modes in convolution. This condition is feasible in computer vision since images are usually of even sizes.

We consider~\eqref{eq:conv-dft} and study the padding mode of $\YY$ when $\XX$ and $\WW$ is symmetrically padded. The transition of the padding modes are listed in Table~\ref{tab:transition}.
\begin{itemize}
    \item We start from an example with $\WSWS(\xx)$, and $\WSWS(\ww)$. It is easy to derive that the output is WSWS-padded. Given an arbitrary output feature map $\yy$, we can pad it as $\WSWS(\yy)$ and obtain $\WSWS(\xx)$ via \eqref{eq:invert-freq}. Since $\WW^{\WSWS}$ is non-zero in general, the inversion is valid. Same inversion can also be achieved for $\xx$ padded in other modes with $\WSWS(\ww)$.
    \item For the case of $\WAWA(\xx)$ and $\WAWA(\ww)$, we get $W^{\WAWA}_0 \equiv W^{\WAWA}_{N/2} \equiv 0$, which make it ill-posed to recover $X^{\WAWA}_0$ and $X^{\WAWA}_{N/2}$ from the output. Fortunately, based on the property of $\WAWA(x)$ we priorly know that $X^{\WAWA}_{N/2} \equiv X^{\WAWA}_{0} \equiv 0$. In addition, it can be derived that the output of the padded convolution in this case is always ZSZS and this form can be obtained by padding arbitrary $\yy$ signal. Thus we are able to recover $\xx$ from arbitrary output feature $\yy$
    \item For the case of $\WAWA(\xx)$ and $\HSHS(\ww)$, we get $\HAHA(\yy)$ as output with $Y^\HAHA_0=Y^\HAHA_{N/2}=0$. However, this does not always hold for arbitrary $\HAHA(\yy)$. Thus we cannot recover $\xx$ from arbitrary $\yy$.
    \item For the case of $\WSWS(\xx)$ and $\HSHS(\ww)$, the inversion is invalid since $W_{N/2} = 0$ and $X_{N / 2} \neq 0$, making it ill-posed to recover $X_{N / 2}$.
\end{itemize}

\subsection{Invertible Convolution}

With the above derivation, we discover multiple forms of invertible convolution with symmetric padding for 1D signals. Different from~\cite{karami2019symmetric} where $\ww$ is restricted to be symmetric, we show that anti-symmetric $\ww$ can also be used to form invertible convolution. 

Alg.~\ref{alg:forward} and \ref{alg:inverse} shows the pseudo-code of the forward and inverse pass of the proposed invertible convolution.
In practice, when defining such invertible convolution, one selects the padding modes for input $\xx$ and $\ww$ from Table~\ref{tab:transition} and pads the signals according to Fig.~\ref{fig:pad-mode}. 
In the forward pass, we implement the symmetrically padded convolution as an equivalent form of circularly padded convolution on symmetrically padded signals. The circularly padded convolution is generally accessible in modern deep learning frameworks and supports automatic gradient computation. In the inverse pass, we  implement \eqref{eq:invert-freq}.
Note that $\ww$ should also be padded to have the same size with $\xx$ so that their DFT can be properly computed. 
In deep learning training frameworks, $\ww$ is denoted as a small parameter tensor. We first pad $\ww$ in the selected padding mode and then resized the padded kernel by filling $0$ to the intended size. The resized kernel has the same padding mode as required.

\begin{algorithm}
\KwIn{$\xx$, $\ww$, $\xpad \in \{ \WSWS, \WAWA, \HSHS, \HAHA, \ZSZS \}$, $\wpad \in \{ \WSWS, \WAWA \}$}
\KwOut{$\yy$, $\ypad$}

Determine $\ypad$ via Table~\ref{tab:transition}\;

$\yy' \gets \mathsf{circular\_conv}(\xpad(\xx), \wpad(\ww))$\;
$\yy \gets \ypad^{-1}(\yy')$
\caption{Forward pass of invertible symmetrically padded convolution}
\label{alg:forward}
\end{algorithm}

\begin{algorithm}
\KwIn{$\yy$, $\ww$, $\ypad \in \{ \WSWS, \WAWA, \HSHS, \HAHA, \ZSZS \}$, $\wpad \in \{ \WSWS, \WAWA \}$}
\KwOut{$\yy$, $\ypad$}

Determine $\xpad$ via Table~\ref{tab:transition}\;
$\YY^\ypad \gets \mathcal{F}(\ypad(\yy))$, $\WW^\wpad \gets \mathcal{F}(\wpad(\ww))$\;

$\XX^\xpad \gets \mathsf{element\_inverse}(\WW^\wpad) \circ \XX^\xpad$\;
$\xx' \gets \mathcal{F}^{-1}(\XX^\xpad)$\;
$\xx \gets \xpad^{-1}(\xx')$
\caption{Inverse pass of invertible symmetrically padded convolution}
\label{alg:inverse}
\end{algorithm}

The symmetric padding in \cite{karami2019symmetric} is a special case of Table~\ref{tab:transition} (row \#4) with $\xx$ and $\ww$ both padded as WSWS. A padded kernel $\WSWS(\ww)$ can be interpreted as a kernel with restricted degree of freedom, i.e., the symmetric points must have equal values. This restricts the flexibliity of the kernel and make it impossible to learn the anti-symmetric components in a kernel.
To increase the capacity to learn anti-symmetric kernels, we can additionally extend the symmetrically padded convolution with parallel anti-symmetrically padded convolution (row \#12 or \#15).

The extension of Table~\ref{tab:transition} is a nature in that the padding modes of the two dimensions follows the 1D transition relationship independently. Take row \#4 and \#12 as two selected modes for symmetric and anti-symmetric padding. There exist 4 different combinations for the 2D padding modes. Note that by decomposing symmetrically and anti-symmetrically on two dimensions, a 2D signal can be decomposed into 4 components. These components corresponds to the 4 padding modes respectively.

\section{Implementation}
\subsection{Implementation Details}

In the implementation of the conventional convolution, a kernel is expressed as a small (e.g., $3 \times 3$) learnable patch. To enable \eqref{eq:conv-dft}, we need to pad and resize a kernel $\ww$ to the same size as the input signal $\xx$. In our implementation, we convert a conventional $\ww$ to $\ww + \mathsf{flip}(\ww)$ for the symmetric case and $\ww - \mathsf{flip}(\ww)$ for the anti-symmetric case to achieve a equivalently padded kernel. This new kernel can be then fed to the conventional convolution operator to realize our proposed invertible convolution.

\subsection{Source code}
The PyTorch source code can be accessed at \url{https://github.com/prclibo/iconv_dft}.

\subsection{Numerical Stability}
We notice that when stacking multiple invertible layers as a network, the numerical error of the inverse pass is accumulated. Executing the computation in double precision helps alleviate the numerical error.

\section{Conclusions}
We comprehensively analyze several different symmetric and anti-symmetric padding modes for the convolution operation and show that multiple cases exist where the convolution can be inverted. We consider this derivation as a contribution to the research of invertible networks and generative models.

\section{Acknowledgement}
Appreciation goes to Marc Finzi for his kind help in this research.

\bibliographystyle{plain}
\bibliography{sample}

\end{document}